\documentclass[10pt,twocolumn,letterpaper]{article}

\usepackage[algorithms]{wacv}  %

\usepackage[T1]{fontenc}
\usepackage[accsupp]{axessibility}  %
\usepackage{multirow}
\usepackage{wrapfig}
\usepackage{placeins}    %
\usepackage{dblfloatfix} %

\definecolor{wacvblue}{rgb}{0.21,0.49,0.74}
\usepackage[breaklinks,colorlinks,allcolors=wacvblue]{hyperref}

\def\wacvPaperID{688}
\def\confName{WACV}
\def\confYear{2027}

\newcommand{\method}{\textsc{RPA}}

\title{Scalable Black-Box Model Attribution for Images %

}

\author{Asaf Livne\\
Tel Aviv University, Israel
\and
Amir Jevnisek\\
Tel Aviv University, Israel
\and
Shai Avidan\\
Tel Aviv University, Israel
}

\begin{document}
\maketitle

\begin{abstract}
The rapid proliferation of generative models raises the model attribution problem: given only an image, can we determine which model produced it? We propose \method{} (Raw-Patch Attribution), a lightweight CNN that solves this problem in a strict black box setting. The CNN operates on multiple image patches to handle varying image size and improve accuracy. It attributes more models at higher accuracy than prior work, reaching 98.9\% on 25-class DRAGON and 95.0\% on 27-class OpenFake; runs in a few milliseconds at a cost nearly independent of candidate-set size; and remains robust to transformations encountered in the wild. Beyond closed-set attribution, its learned representation serves as a reusable fingerprint backbone: it supports open-set rejection and few-shot enrollment, recovers model-family structure without lineage supervision, and groups images from unseen generators by source. Controlled ablations and causal perturbations show that its attribution decisions are driven by a spatially local, low-level signal that behaves as a generator-specific fingerprint.
\end{abstract}
\section{Introduction}
\label{sec:introduction}

\begin{figure}[t]
    \centering
    \includegraphics[width=\linewidth]{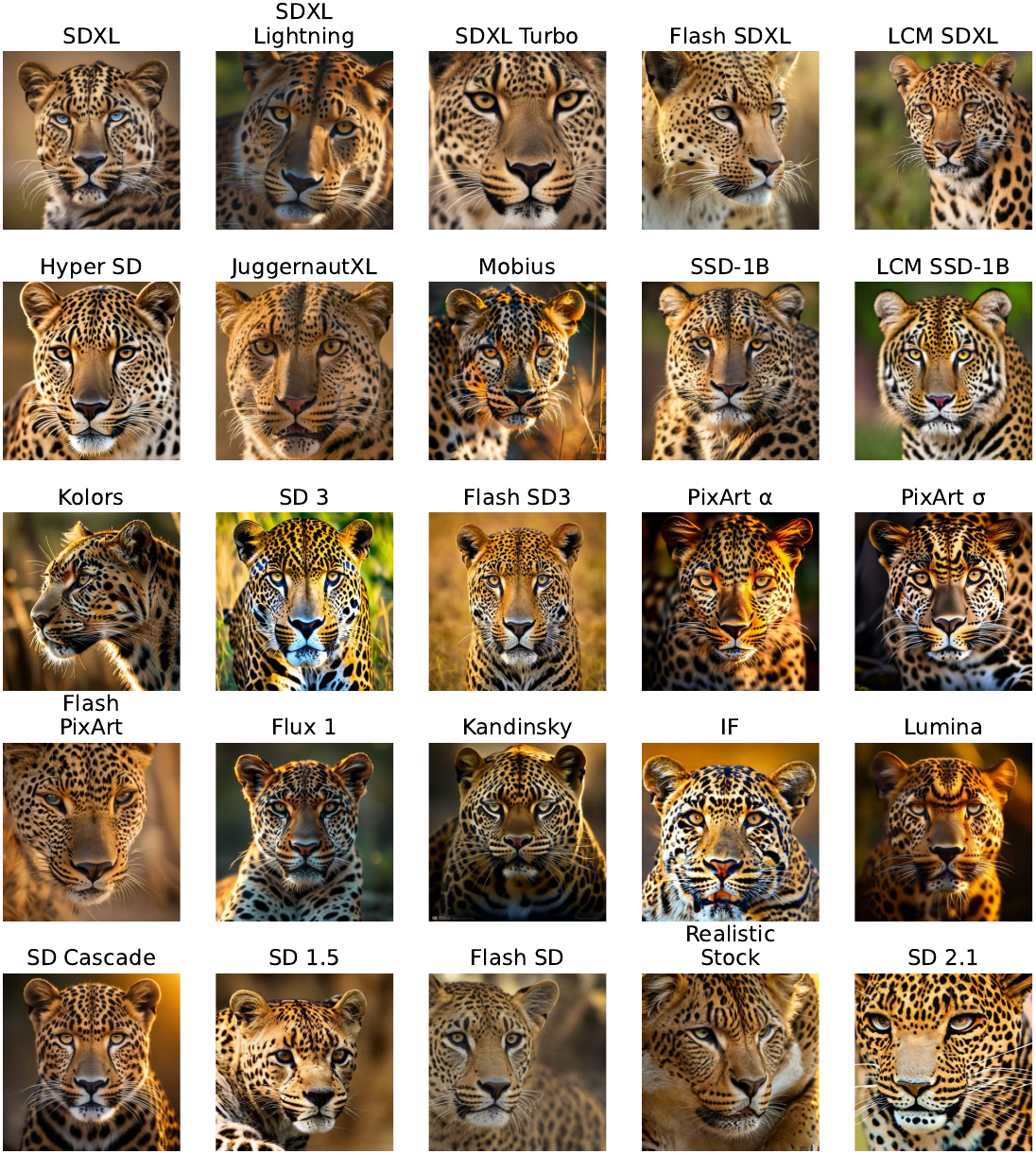}
    \caption{%
        \textbf{The same prompt rendered by 25 generators
        (DRAGON~\citep{dragon2024}).} The outputs are visually similar,
        yet a small CNN trained on raw RGB patches attributes each image
        to its source model at $98.9\%$ accuracy.
    }
    \label{fig:teaser}
\end{figure}

Text-to-image diffusion models~\citep{rombach2022ldm,podell2023sdxl,esser2024sd3}
are rapidly becoming the infrastructure
for creative work, media production, and visual communication. As their
outputs spread, tracing an image back to its source model, a task known
as \emph{model attribution}~\citep{yu2019attribution,marra2019ganfingerprints,song2024manifpt},
grows increasingly valuable. Its current uses are mostly forensic, such
as IP protection and content provenance~\citep{fernandez2023stable,wen2023treering},
but they can reach further, to
recommending content or collecting royalties on a model's outputs. The ecosystem that
makes attribution pressing also makes it hard: a few
dozen to a few hundred base models anchor the field, and each spawns a
long tail of fine-tunes, merges, and distillations, so open
model-sharing platforms already host hundreds of thousands of distinct
checkpoints~\citep{horwitz2025atlas}, with new ones appearing daily.

Despite growing interest, existing attribution methods remain far from
practical deployment. Current approaches suffer from one or more
limitations: they require access to model internals such as weights,
autoencoders, or prompts that are unavailable in realistic
scenarios~\citep{yang2024fingerprinting,ricker2024aeroblade,sha2023defake};
they achieve only coarse-grained discrimination, failing to distinguish
models that share an autoencoder or belong to the same architectural
family~\citep{wang2025aedr}; they are fragile under common image
degradations such as JPEG compression and
resizing~\citep{gragnaniello2021gan}; or they impose
substantial computational cost that limits
scalability~\citep{wang2023dire,wang2024latenttracer}.

Many of these methods trace their lineage to synthetic-image
\emph{detection}. Spectral~\citep{frank2020leveraging,corvi2023diffusion},
reconstruction-error~\citep{wang2023dire,ricker2024aeroblade}, and
pretrained-encoder~\citep{ojha2023universal} pipelines were developed
first to separate real images from generated ones, and only later
carried over to attribution, in several cases as a by-product of the
detector itself~\citep{sinitsa2024deep,sha2023defake}.
Detection and attribution, however, seek different evidence.
Detection looks for cues shared across generated images that separate
them from real ones. As text-to-image models improve, semantic and
perceptual evidence for this boundary becomes increasingly unreliable.
Attribution can instead rely on source-specific residues left by the
generation pipeline: like sensor pattern noise in camera forensics,
these traces need not be perceptible or shared across generators, only
stable within a source and separable between sources. Closed-set
attribution therefore does not require a representation that generalizes
a real--fake boundary to unseen generators. Nevertheless, we find such
generalization to be an emergent property: training a classifier to
separate many generators induces a feature space that naturally extends
beyond the training set.

We therefore revisit supervised classification as a means of learning a
generator's fingerprint directly from its
images~\citep{yu2019attribution,yang2022dnadet,bondi2017camera}.
Our method, \method{} (Raw-Patch Attribution), splits an image into
$64{\times}64$ RGB patches, classifies each patch with a compact CNN,
and aggregates the per-patch logits into an image-level prediction.
Operating across dozens of modern generators at once, we show that it
attributes each image to its source despite the
visual similarity in Figure~\ref{fig:teaser}.
It reaches \textbf{98.9\%} on 25-class
DRAGON~\citep{dragon2024} and \textbf{95.0\%} on 27-class
OpenFake~\citep{openfake2024}, operating strictly \emph{black-box} from
the output image alone, with no access to weights, autoencoders, or
prompts. To the best of our knowledge, no prior method attributes across
this many models at this level of accuracy. A corruption-augmented
variant withstands the JPEG compression, blur, resizing, and cropping
that defeat prior methods. Both training and inference are inexpensive:
the model is compact, trains in a short run, and attributes each image
in a few milliseconds, at a cost effectively independent of the number
of candidate models.

Crucially, the classifier learns more than a fixed set of
closed-set decision boundaries. Although trained only to separate known
generators, its penultimate representation forms an emergent fingerprint
space that captures relationships among generative models and
generalizes beyond the training classes.
The same network also supports three further tasks without relearning the
representation. In \textbf{open-set attribution}, the network successfully
attributes images from known models while reliably flagging inputs from unseen
generators as unknown, using only the classifier's own confidence. For
\textbf{few-shot adaptation}, a new generator is admitted by fitting only a
linear head rather than a full retrain, with few-shot transfer to a different
benchmark completing in seconds. And for \textbf{model discovery}, its
penultimate-layer features inherently organize the generative ecosystem,
allowing us to recover the lineage of trained models via hierarchical clustering
and to faithfully group images from unseen generators by their true source.

Our contributions are:
\begin{itemize}[nosep,leftmargin=*]
    \item \textbf{Accurate black-box attribution.} A lightweight CNN on raw
    image patches provides accurate ($98.9\%$ on DRAGON, $95.0\%$ on
    OpenFake) and efficient black-box attribution that is robust to common
    image degradations, with single-pass inference that is highly scalable.
    \item \textbf{Open-set attribution and few-shot adaptation.} The network 
    reliably attributes known models while rejecting unseen ones, and admits 
    new generators by fitting only a linear head, with few-shot transfer 
    completing in seconds.
    \item \textbf{Model discovery.} The learned features naturally encode 
    relationships between generators, allowing us to recover the family lineage 
    of known models and faithfully group images from unseen generators.
\end{itemize}
\section{Background}
\label{sec:related_work}

\paragraph{Evidence and tasks.}
Tracing generative outputs relies on two types of evidence.
\emph{Watermarks} are signals actively embedded during
generation~\citep{wen2023treering,fernandez2023stable,yang2024gaussianshading},
which requires provider cooperation and often degrades generation
quality; in practice, most images in the wild carry no watermark.
\emph{Fingerprints} are intrinsic traces involuntarily left by a
model's architecture and generation
pipeline~\citep{marra2019ganfingerprints,yu2019attribution,song2024manifpt}, the
generative analogue of sensor pattern noise in camera
forensics~\citep{lukas2006detecting}, where patch-level CNN
classifiers with aggregated per-patch decisions are long
established~\citep{bondi2017camera}.
These serve two forensic tasks: \emph{detection} (binary
real-vs.-fake classification) and \emph{source attribution}
(identifying \emph{which} model produced a given image).
Attribution is a distinct task and the focus of this work.

\paragraph{Problem settings.}
Attribution methods sit along two axes. The first is the
candidate set: \emph{closed-set} methods assume the source is one
of a fixed set of known generators~\citep{yu2019attribution,yang2022dnadet,xu2024originattrib}, with abundant labels per generator or only a handful in the \emph{few-shot} regime~\citep{yuan2024occclip,wang2026lida}, while \emph{open-set} methods
drop this assumption and, beyond attributing images from the known
generators, must either flag an unseen source as
unknown~\citep{yang2023pose,sun2023cpl,fang2023openset,laszkiewicz2024single} or discover and
group new sources without labels~\citep{girish2021openworld,cozzolino2025forensic}.
The second is access: \emph{white-box} methods need
model internals such as weights~\citep{yang2024fingerprinting} or
a candidate autoencoder~\citep{ricker2024aeroblade,wang2025aedr},
whereas \emph{black-box} methods use only the output image. This
work operates in the strictest black-box setting, using the image
alone without prompt or seed, and performs both closed-set and
open-set attribution.

\label{sec:rw_attribution}
\paragraph{Existing methods.}
Binary detection has matured: a CNN trained on one generator
generalizes to others~\citep{wang2020cnndetection}, frozen CLIP
features give a strong cross-generator probe~\citep{ojha2023universal},
and recent work pushes generalization to unseen generators through
data diversity~\citep{cozzolino2025community,guillaro2025bfree} and
MLLM-based explanation~\citep{zhou2025aigiholmes,tan2025forenx}.
Attribution draws on the same methods, which fall into four families.
\emph{Reconstruction-based} methods score an image by how well each
candidate reconstructs it~\citep{wang2023dire,ricker2024aeroblade,wang2023origin,wang2024latenttracer,wang2025aedr},
inherently requiring white-box access.
\emph{Pixel-frequency} methods analyze spectral artifacts in the Fourier
or DCT domain~\citep{frank2020leveraging,corvi2023diffusion,sinitsa2024deep,karageorgiou2025spai}, including CNNs trained directly on the
spectrum~\citep{zhang2019autogan}.
\emph{Encoder-based} methods adapt pretrained encoders, such as CLIP,
for attribution~\citep{sha2023defake,yuan2024occclip,cioni2024clip}.
\emph{Discriminative} classifiers learn the source directly from the
image, the oldest approach and the closest to
ours~\citep{yu2019attribution,yang2022dnadet}, now extended to
end-to-end~\citep{xu2024originattrib} and open-set
forms~\citep{yang2023pose,sun2023cpl}.

\paragraph{Model discovery.}
A growing body of work asks not just which model produced an image,
but how to discover and map relationships between models, including
unknown ones. While lineage mapping typically operates in weight space
to chart known checkpoints~\citep{yu2024neurallineage,yu2025neuralphylogeny,horwitz2025mothr,horwitz2025atlas},
discovery extends to identifying and placing unseen sources. Closer to
us, ManiFPT~\citep{song2024manifpt} shows fingerprint clusters reflect
architectural design, and \citet{sinitsa2024deep} trace relationships
through shared spectral artifacts to group models. We discover
and place models based on genealogy recovered from output images
alone, without model weights, as an emergent property of the
attribution features.
\section{Method}
\label{sec:method}

\begin{figure}[t]
    \centering
    \includegraphics[width=\linewidth]{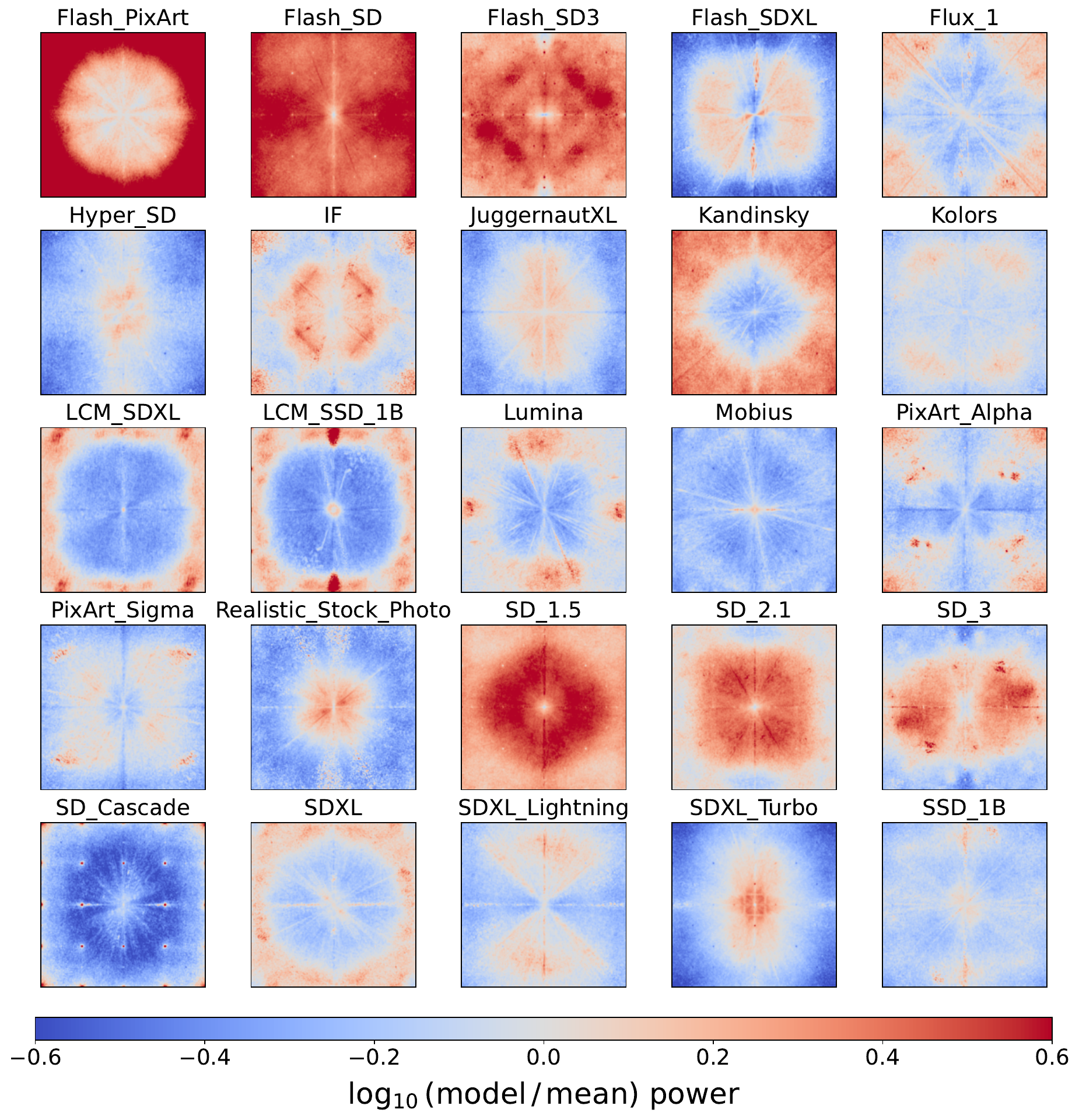}
    \caption{\textbf{Each generator leaves a distinct, stable spectral signature.} Per-generator 2D patch power spectra for the 25 DRAGON~\citep{dragon2024} generators, each plotted as a log-ratio to the across-model mean.}
    \label{fig:spectral_signatures}
\end{figure}

Generative pipelines embed persistent, highly separable artifacts in the frequency domain. As illustrated in Figure~\ref{fig:spectral_signatures}, the 2D power spectra of patches from 25 DRAGON~\citep{dragon2024} generators reveal distinct, stable signatures for each source; this analysis serves only to expose the artifacts and is not used for attribution (\S\ref{sec:exp_ablation}). The low-level, local nature of these traces suggests that a lightweight CNN operating on raw patches is sufficient to capture them. Accordingly, our method splits the image into patches, classifies each patch with this CNN, and aggregates the per-patch logits into an image-level prediction.

\subsection{Patch Division}
We split the input image $x \in \mathbb{R}^{H \times W \times 3}$ into $64 \times 64$ patches:
\begin{equation}\label{eq:patchify}
    p_i = \Pi_i(x) \in \mathbb{R}^{64 \times 64 \times 3}.
\end{equation}
When $H$ or $W$ is not a multiple of $64$, edge patches overlap so that every pixel is covered. Patching makes the classifier resolution-invariant: the same model runs on inputs from $64^2$ to $4096^2$ without retraining. It yields several patches per image, which are aggregated at inference, and a smaller input simply produces fewer patches.

\subsection{Per-Patch Classification}
Each patch $p_i$ is classified from its raw pixels by a CNN $f$:
\begin{equation}\label{eq:perpatch}
    \ell_i = f(p_i) \in \mathbb{R}^{C},
\end{equation}
where $\ell_i$ are class logits and $C$ is the number of candidate generators. Each patch inherits its image's label, and $f$ is trained to minimize the cross-entropy between the softmax $\sigma(\ell_i)$ and that label. We deliberately keep $f$ small, a compact convolutional network of about $6$M parameters, as the low-level fingerprint needs no heavy backbone; the full architecture and training recipe are in the supplementary material.

Because $f$ produces logits for all $C$ candidate generators in a single forward pass, the per-image cost is almost  independent of the number of candidates: admitting a new generator only widens the final layer and leaves inference almost unchanged, effectively $O(1)$ in $C$: the head stays under $1\%$ of inference cost up to $C{\approx}6{,}000$ (supplementary). Reconstruction-based attribution instead scores a query by inverting or reconstructing it through each candidate in turn, and therefore scales as $O(C)$. This gap widens as model registries grow; we quantify it against white-box baselines in \S\ref{sec:exp_whitebox}.

\subsection{Multi-Patch Aggregation}
\label{sec:aggregation}

Each patch provides local evidence about the image's source, and
combining these predictions improves image-level attribution.

The image-level prediction is a weighted average of the per-patch
logits:
\begin{equation}
    \hat{y} = \arg\max_k \;\Bigl(\sum_{i=1}^{P} w_i \,
    f(p_i)\Bigr)_k,
    \label{eq:aggregate}
\end{equation}
where the weight $w_i$ counts each patch inversely to its edge overlap, so
that every pixel contributes equally; with no overlap this reduces to a
plain average.
\section{Experiments}
\label{sec:experiments}

\subsection{Experimental Setup}
\label{sec:exp_setup}

\paragraph{Datasets.}
We evaluate on three benchmarks:
\begin{itemize}[nosep,leftmargin=*]
    \item \textbf{DRAGON}~\citep{dragon2024}: A controlled benchmark of 25 recent generative models, many of them closely related (fine-tunes, distillations, and roughly ten that share a single SDXL autoencoder). All models are prompted with the same captions and use default generation parameters, eliminating semantic and configuration bias. Most models produce $1024{\times}1024$ images. We use the ``Regular'' split (750 train / 250 val / 400 test per class).
    \item \textbf{OpenFake}~\citep{openfake2024}: An uncontrolled, real-world benchmark whose images were \emph{collected} in the wild rather than generated by the benchmark authors, covering 27 open-source and commercial generators (including closed-source models) at varied resolutions (512--2048\,px), with heterogeneous prompts and generation settings. We use the same per-class split as DRAGON (750 train / 250 val / 400 test).
    \item \textbf{AEDR}~\citep{wang2025aedr}: An eight-model benchmark of open latent-diffusion generators, used for the white-box comparison (\S\ref{sec:exp_whitebox}) because neither main benchmark exposes a distinct, accessible autoencoder per candidate model. We reconstruct it from AEDR's released prompts and generation pipeline, generating images for its eight open models and splitting them per model into disjoint partitions ($350/50/100$ train/val/test); we evaluate on the $800$ held-out test images ($100$ per model).
\end{itemize}
Robustness and the ablations run on DRAGON, whose controlled conditions
isolate the single factor under test; open-set attribution, lineage, and
discovery use OpenFake for its in-the-wild model diversity; adaptation
spans both benchmarks and GenImage~\citep{zhu2023genimage}; and the
white-box comparison uses AEDR.

\paragraph{Evaluation.}
The two settings differ in granularity. Black-box attribution is
\emph{multi-class}: we report top-1 image-level accuracy over every
generator in the benchmark (25 on DRAGON, 27 on OpenFake). The
white-box comparison is \emph{binary}: following AEDR's native pairwise
protocol, each image is assigned to one of two candidate models, scored
over the $27$ pairs AEDR reports (the $28$ model pairs minus the
near-identical SD2-base--SD2.1 pair, which AEDR itself excludes). Unless
stated otherwise, accuracy is on the held-out test split; the open-set
and clustering experiments add metrics defined in their respective
subsections.

\subsection{Closed-Set Attribution}
\label{sec:exp_blackbox}

\begin{figure*}[!t]
    \centering
    \includegraphics[width=\linewidth]{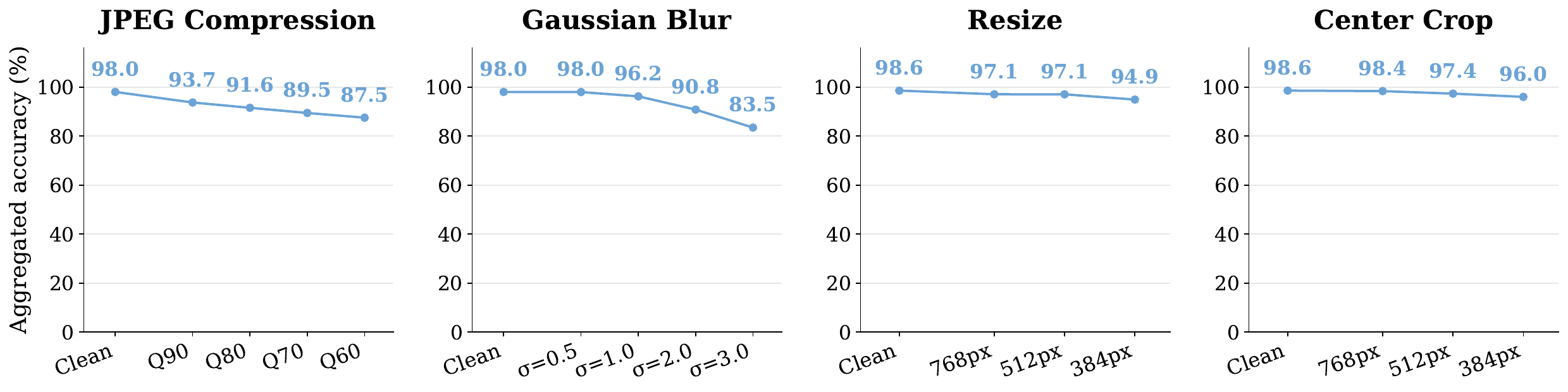}
    \caption{%
        \textbf{Attribution is robust to common image degradations.}
        Accuracy of a single model across four degradation families
        (DRAGON 25-class), starting from a clean 98.0\%; heavy
        blur is the hardest condition. Resize and crop are measured on the
        twenty generators that natively render at $1024{\times}1024$, hence
        their higher clean baseline.
    }
    \label{fig:robustness}
\end{figure*}

\paragraph{Black-box comparison.}
Our method achieves \textbf{98.9\%} accuracy on 25-class DRAGON and
\textbf{95.0\%} on 27-class OpenFake
(Table~\ref{tab:blackbox}), against
$62.0\%$ for the strongest baseline, a ${\sim}$35$\times$ higher error
rate. Errors concentrate within model families; the confusion matrices
are in the supplementary. That baseline, DE-FAKE~\citep{sha2023defake}, is the sole
attribution method DRAGON~\citep{dragon2024} itself evaluates, and it
additionally requires the generating text prompt.
The others reuse a frozen representation rather than learning
one: USIA~\citep{cioni2024clip} probes CLIP features,
LIDA~\citep{wang2026lida} retrieves against the training set as gallery,
and OCC-CLIP~\citep{yuan2024occclip} takes the argmax over one-vs-real
detectors. We re-measured every method's inference time on our own hardware,
where ours is the fastest of the five.
EfficientFormer~\citep{xu2024originattrib} is closest to ours in spirit
but releases neither code nor data, so we quote rather than tabulate it:
$91.0\%$ on 13 classes of private data with ${\sim}4\times$ more images per
class.

\begin{table}[!t]
\centering
\caption{%
    \textbf{Black-box setting.}
    From the image alone, a compact CNN attains the highest accuracy
    at the largest class count, at the lowest inference cost.
}
\label{tab:blackbox}
\footnotesize
\setlength{\tabcolsep}{2pt}
\begin{tabular}{@{}lrrcc@{}}
\toprule
\textbf{Method} & \textbf{Params\,(M)} & \textbf{Infer.\,(ms)} & \textbf{DRAGON\,(25)} & \textbf{OpenFake\,(27)} \\
\midrule
DE-FAKE~\citep{sha2023defake}        & 151 & $9.96_{\pm1.31}$ & 62.0 & 57.3 \\
USIA~\citep{cioni2024clip} & 427 & $9.61_{\pm1.01}$ & 52.1 & 50.1 \\
LIDA~\citep{wang2026lida}  & 23.5 & $23.6_{\pm12.5}$ & 24.0 & 17.7 \\
OCC-CLIP~\citep{yuan2024occclip}     & 151 & $7.87_{\pm1.12}$ &  8.6 & 15.8 \\
\midrule
\textbf{Ours} & 5.9 & $\mathbf{2.79_{\pm0.04}}$ & \textbf{98.9} & \textbf{95.0} \\
\bottomrule
\end{tabular}
\end{table}

\paragraph{White-box comparison.}
\label{sec:exp_whitebox}
On AEDR's eight-model benchmark our
classifier reaches \textbf{99.5\%} mean pairwise accuracy
(Table~\ref{tab:whitebox}), despite operating under
black-box access: it sees only the output image and never a candidate
autoencoder.
Our method also reaches \textbf{98.8\%} in the harder 8-way single-label
setting; reconstruction-based pairwise scoring extends to this
regime only with difficulty, as it must rank each image against all
eight candidates at once rather than separate a single pair.
Beyond accuracy, our inference-time advantage over these methods
grows with the number of candidate models $C$. Specifically, our 
inference time remains a constant 2.8\,ms, whereas AEDR's cost 
scales linearly, rising from 0.27\,s to 1.87\,s as $C$ grows from 
2 to 8.

\begin{table}[!t]
\centering
\caption{%
    \textbf{White-box setting.}
    Mean pairwise accuracy and per-image inference time on AEDR's
    eight-model benchmark. Our method outperforms the alternatives while being two orders of magnitude faster and strictly black-box.
}
\label{tab:whitebox}
\footnotesize
\setlength{\tabcolsep}{5pt}
\begin{tabular}{@{}lccc@{}}
\toprule
\textbf{Method} & \textbf{Access} & \textbf{Infer.\,(s)} & \textbf{Acc.\,(\%)} \\
\midrule
LatentTracer~\citep{wang2024latenttracer} & Model weights & $24.06_{\pm0.01}$  & 70.3 \\
AEDR~\citep{wang2025aedr}                 & VAE weights & $0.267_{\pm0.003}$  & 95.1 \\
\midrule
\textbf{Ours} & \textbf{Image only} & $\mathbf{0.00279_{\pm0.00004}}$ & \textbf{99.5} \\
\bottomrule
\end{tabular}

\vspace{2pt}
\begin{minipage}{\columnwidth}
\footnotesize Baseline accuracies are taken from
AEDR~\citep{wang2025aedr}; all inference times are re-measured on our
hardware.
\end{minipage}
\end{table}

\subsection{Robustness}
\label{sec:exp_robust}

Real-world images undergo lossy transformations before analysis.
We test whether a single model maintains attribution accuracy
under four common degradations: JPEG compression, Gaussian blur,
resizing, and center cropping.
We train one model on all 25 DRAGON generators with
stochastic augmentation (JPEG~$Q{\in}[60{,}90]$,
blur~$\sigma{\in}[0.1{,}4.0]$, resize~${\in}[384{,}768]\,$px)
alongside clean copies. Corruptions are never composed: each
training image is either left clean or augmented with a single
corruption (full ranges in the supplement). No crop
augmentation is used. Resize and crop target an absolute
resolution, which changes meaning for the five generators that render
below $1024{\times}1024$; those two families are evaluated on the other
twenty, with predictions still ranging over all 25 classes.

Figure~\ref{fig:robustness} reports accuracy across 15~conditions.
The model degrades gracefully on every family, and it comes at a
manageable cost on clean data: the augmented model scores $98.0\%$ against
$98.9\%$ for a clean-trained one.

\subsection{Open-Set Attribution}
\label{sec:exp_openset}

A deployed attributor meets generators absent from its training set.
We partition the 27 OpenFake generators into 17~\emph{known} and
10~\emph{unseen}, repeat over five random draws, and train a 17-class
CNN on the known generators ($95.8\%{\pm}1.3$ closed-set accuracy).
The classifier's own confidence successfully flags out-of-set images: scoring each
test image by its top-class margin, with no calibration set and no
auxiliary outlier data, separates unseen from known generators at an open-set
classification rate of \textbf{AU-OSCR~0.875}${\pm}0.037$, averaged over the five draws. The score
tracks the difficulty of the drawn split rather than the number of
known anchors: the best draw reaches AU-OSCR~0.918, on
par with the single $0.913$ AU-OSCR reported by optimization-based
detection~\citep{cozzolino2025forensic}, at ${\sim}3{\times}$ the
model count (27 vs.\ 9) and ${\sim}3000{\times}$ lower inference
cost (${\sim}2.7$\,ms vs.\ ${\sim}9$\,s).
Rejection is weakest for finetune variants that inherit a known family's fingerprint,
which the classifier confidently assigns to a known sibling. Full
per-generator recall and OSCR curves are in the supplementary.

\subsection{Adapting to New Generators}
\label{sec:exp_adapt}

\begin{figure*}[!b]
    \centering
    \includegraphics[width=0.96\linewidth]{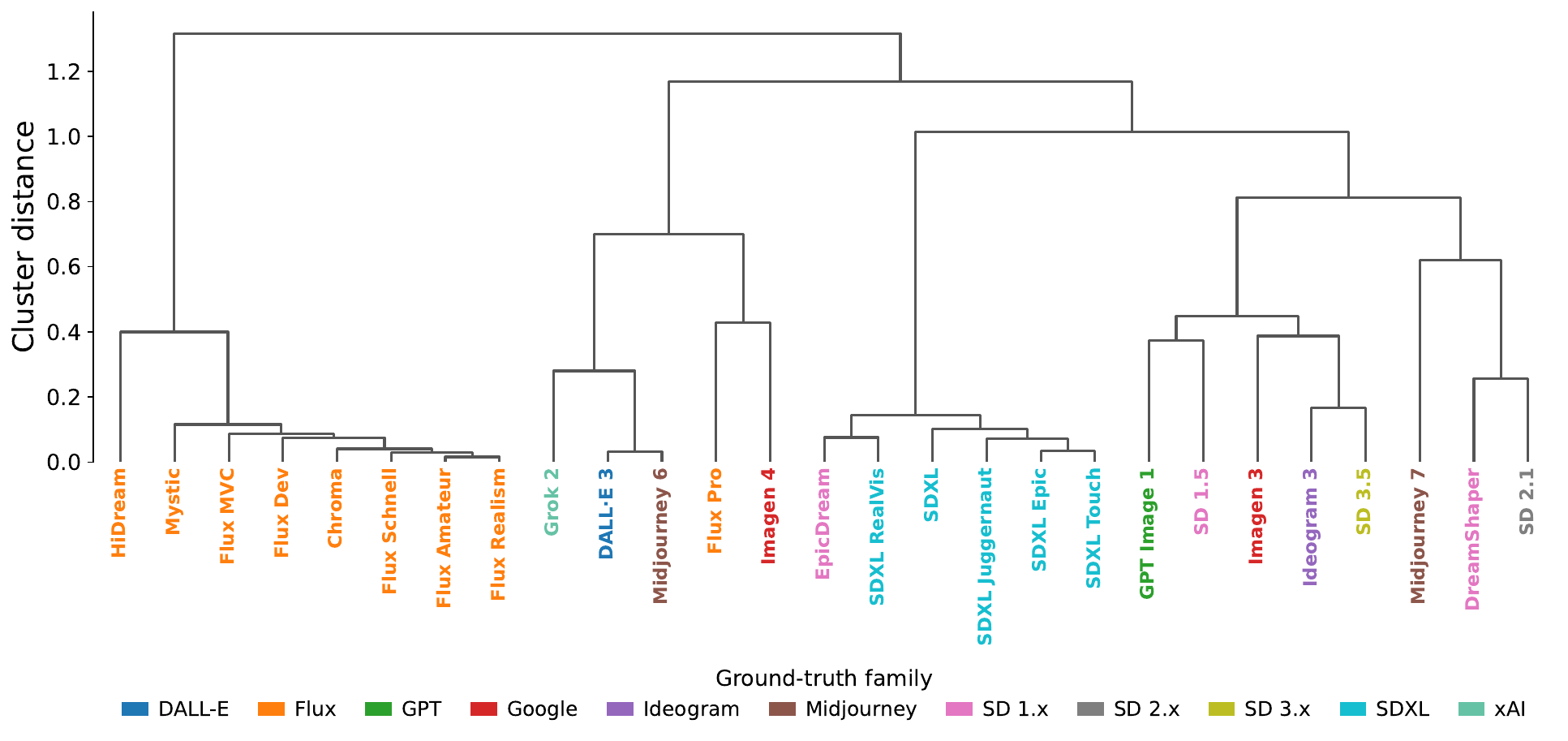}
    \caption{%
        \textbf{Generators cluster by architectural family without
        supervision.} Hierarchical clustering of the per-generator
        CNN feature vectors (27 OpenFake generators); leaf colors mark
        ground-truth families (shared base weights or autoencoder).
        The Flux family (including HiDream, Mystic, and Chroma) and the
        SDXL variants each form a clean clade.
    }
    \label{fig:dendogram}
\end{figure*}

A deployed attributor must keep pace with generators that appear after
training. We find that the backbone learns a general fingerprint space
rather than a fixed set of class boundaries, so a new generator can be
admitted by freezing the backbone and fitting only a linear head, with
full adaptation completing in under two minutes. We demonstrate this
across three regimes.

\paragraph{Full adaptation.}
To establish the upper bound of accuracy when adapting only the linear head, we
extend the 17-known OpenFake model of \S\ref{sec:exp_openset} to all 27
generators on the full dataset. Freezing its backbone, we fit a fresh 27-way linear head on
cached features, at $0.4\%$ of the parameters. The ten new
generators are learned while performance on the original 17 is largely preserved (Table~\ref{tab:adapt}).

\paragraph{Cross-dataset adaptation.}
The same recipe carries an entire generator set across benchmarks. Freezing a
backbone trained on one benchmark and fitting a head on the full, disjoint
generator set of another, an OpenFake backbone attributes DRAGON's 25 generators
at $96.6\%$ (vs.\ $98.9\%$ for DRAGON's own model). The reverse,
DRAGON\,$\rightarrow$\,OpenFake, reaches $86.8\%$ (vs.\ $95.0\%$): the broader
source yields the more transferable fingerprint, and the gap concentrates on
OpenFake's Flux and SDXL fine-tune families, which a DRAGON backbone has seen
only as single instances. Per-generator breakdowns are in the supplementary.

\paragraph{Few-shot adaptation.}
We test the ability to adapt with very little data to entirely different generators. Freezing an OpenFake-
or DRAGON-trained backbone, we fit a 9-way head on
GenImage~\citep{zhu2023genimage} from a few labels per generator, under the
few-shot protocol of LIDA~\citep{wang2026lida}. This lightweight process takes under four seconds.
Although the source and target datasets share only one overlapping model architecture, the frozen fingerprint space generalizes well across the remaining architectures in GenImage, outperforming the baselines (Table~\ref{tab:fewshot}). Crucially, backbones trained on either source dataset yield comparable accuracy, suggesting the learned representation captures generic structural properties of generative models rather than dataset-specific artifacts.

\begin{table}[!t]
\centering
\caption{%
    \textbf{Head-only adaptation preserves old classes performance while adding
    new ones.} A frozen 17-class backbone with a freshly fit 27-way
    linear head ($28$\,K parameters, ${\sim}2$\,min). \emph{Before} is
    the base model's closed-set accuracy on its 17 classes;
    \emph{after} is post-adaptation. Mean\,$\pm$\,std over five 17/10
    draws.
}
\label{tab:adapt}
\footnotesize
\setlength{\tabcolsep}{6pt}
\begin{tabular}{@{}lcc@{}}
\toprule
\textbf{Eval.\ subset} & \textbf{Before} & \textbf{After} \\
\midrule
Original 17 & $95.8_{\pm1.3}$ & $93.3_{\pm1.8}$ \\
New 10      & ---             & $85.1_{\pm4.2}$ \\
All 27      & ---             & $90.3_{\pm0.5}$ \\
\bottomrule
\end{tabular}
\end{table}

\begin{table}[!t]
\centering
\caption{%
    \textbf{Few-shot attribution on unseen generators.} Top-1 accuracy across nine GenImage~\citep{zhu2023genimage} generators with $N$ labeled images per class, fitting only a linear head on a frozen backbone against LIDA's baselines under their protocol~\citep{wang2026lida}. Mean\,$\pm$\,std over five draws.
}
\label{tab:fewshot}
\footnotesize
\setlength{\tabcolsep}{5pt}
\begin{tabular}{@{}lcc@{}}
\toprule
\textbf{Method} & \textbf{1-shot} & \textbf{10-shot} \\
\midrule
ResNet~\citep{he2016resnet}  & 17.4 & 21.4 \\
DIRE~\citep{wang2023dire}    & 14.3 & 17.2 \\
ESSP~\citep{chen2024ssp}     & 17.0 & 22.4 \\
LIDA~\citep{wang2026lida}    & 40.4 & 54.0 \\
\midrule
Ours (OpenFake backbone)     & $47.7_{\pm3.5}$ & $72.0_{\pm0.8}$ \\
Ours (DRAGON backbone)       & $\mathbf{52.0}_{\pm3.6}$ & $\mathbf{72.4}_{\pm0.9}$ \\
\bottomrule
\end{tabular}

\vspace{4pt}
\footnotesize Baselines as published in LIDA~\citep{wang2026lida}.
\end{table}

\subsection{Ablations}
\label{sec:exp_ablation}

We ablate three design choices, varying one at a time: the input
representation, the classifier head, and the patch size. All use DRAGON-Small,
which provides $75$/$25$/$400$ train/val/test images per class.

\paragraph{Input representation and architecture.}
Vision systems commonly build on a strong pretrained backbone. We
therefore replace the raw patch with the latent of a frozen encoder,
leaving the classifier unchanged: three diffusion autoencoders, two
foundation encoders~\citep{radford2021clip,oquab2025dinov3}, and a $2$D
DCT log-magnitude, each with both a CNN and an MLP head, of which we
report the better. Raw pixels are best (Table~\ref{tab:ablation}a). The
diffusion autoencoders come closest, the semantic encoders trail
furthest, and the fixed DCT basis falls between. No encoder helps; the best performance is achieved when feeding the CNN with the raw pixel data.

\paragraph{Patch size.}
We train one model per patch size from $16$ to $1024$ on the $20$
DRAGON-Small generators that render natively at $1024{\times}1024$
(Table~\ref{tab:ablation}b). Multi-patch accuracy climbs steeply as
patches shrink, plateaus from $128$ to $32$\,px, and dips at $16$, while
single-patch accuracy peaks at $128$\,px: smaller tiles carry no cleaner fingerprint, only more votes.

\begin{table}[t]
\centering
\footnotesize
\setlength{\tabcolsep}{3pt}
\begin{tabular}[t]{@{}llc@{}}
\multicolumn{3}{@{}l@{}}{\textbf{(a) Representation}}\\[2pt]
\toprule
Input & Head & Acc. \\
\midrule
DINOv3          & CNN & 44.9 \\
CLIP            & MLP & 55.6 \\
DCT log-mag     & MLP & 70.9 \\
SDXL            & CNN & 80.6 \\
Flux-dev        & CNN & 88.3 \\
SD-3            & CNN & 89.5 \\
Raw RGB \emph{(ours)} & CNN & \textbf{91.3} \\
\bottomrule
\end{tabular}
\hfill
\begin{tabular}[t]{@{}lcc@{}}
\multicolumn{3}{@{}l@{}}{\textbf{(b) Patch size}}\\[2pt]
\toprule
Size & Single-patch & Multi-patch \\
\midrule
$1024$ & 51.7 & 51.7 \\
$512$  & 61.8 & 73.4 \\
$256$  & 75.1 & 91.3 \\
$128$  & \textbf{88.1} & 98.2 \\
$64$   & 86.9 & \textbf{98.6} \\
$32$   & 81.3 & 98.0 \\
$16$   & 69.8 & 96.3 \\
\bottomrule
\end{tabular}
\caption{%
    \textbf{Ablations on DRAGON-Small.} One factor varied at a time.
    \emph{(a)} each representation (CLIP ViT-L/14, DINOv3 ViT-H/16+, and
    three diffusion autoencoders) is reported with its better head, a CNN
    or an MLP. \emph{(b)} patch size, on the
    $20$ native-$1024$ classes.}
\label{tab:ablation}
\end{table}

\section{Fingerprint Analysis}
\label{sec:fingerprint_analysis}

The penultimate layer of our trained network provides a strong fingerprint that can be utilized for various downstream tasks beyond attribution, sharing the same representation that enables the rapid adaptation shown in \S\ref{sec:exp_adapt}.

\begin{figure}[!t]
    \centering
    \includegraphics[width=\linewidth]{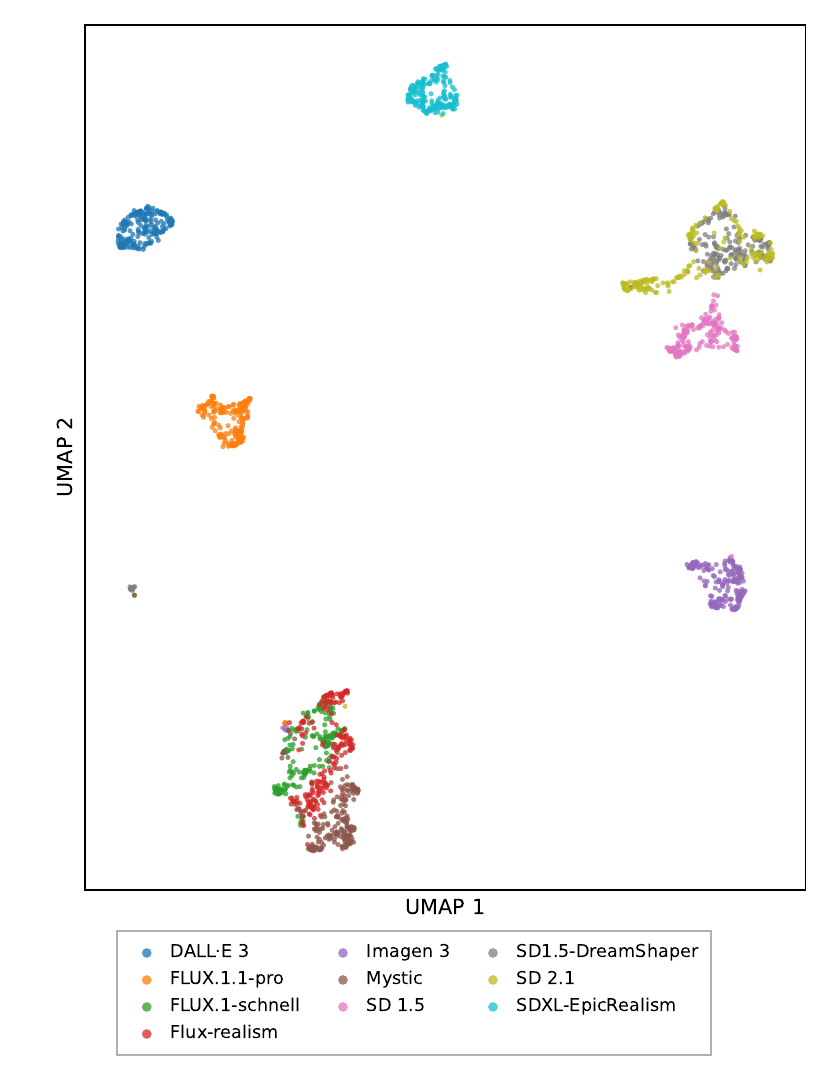}
    \caption{%
        \textbf{Unseen generators separate without supervision.}
        UMAP of penultimate CNN features for the ten held-out
        generators, colored by true source.}
    \label{fig:discovery_umap}
\end{figure}

\begin{figure*}[!t]
  \centering
  \includegraphics[width=0.46\linewidth]{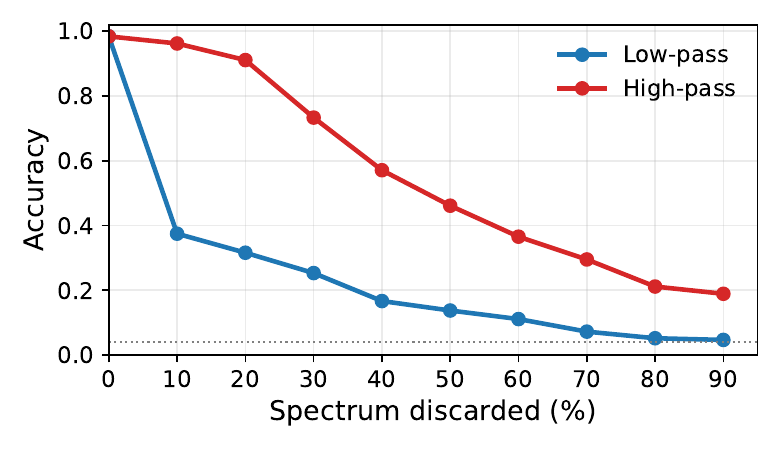}\hfill
  \includegraphics[width=0.46\linewidth]{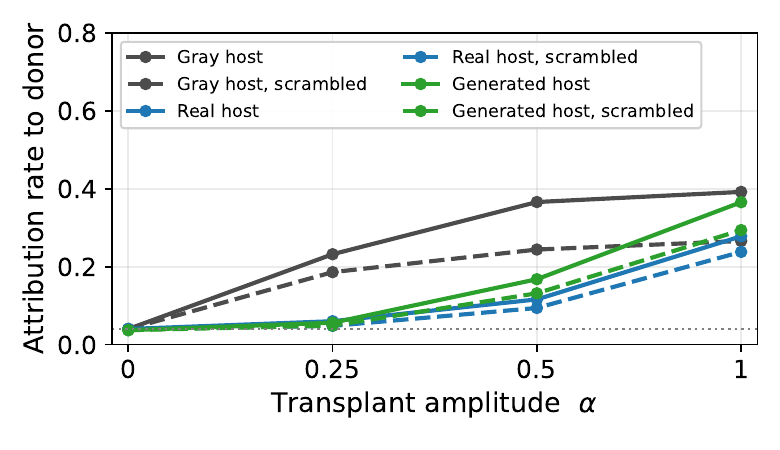}
  \caption{%
    \textbf{The fingerprint is a local, high-frequency texture.}
    \emph{Left:} accuracy under progressive low- or high-pass filtering
    (Butterworth). \emph{Right:} attribution to a donor generator whose
    high band is transplanted onto a real photograph, a flat gray field,
    or another generator's image, against transplant amplitude $\alpha$,
    with the transplanted
    band left intact or scrambled into $8$\,px blocks. Dotted lines mark
    chance.}
  \label{fig:whatseen}
\end{figure*}

\subsection{Lineage Analysis}
\label{sec:analysis_lineage}

A CNN trained purely for model attribution receives no lineage
labels, yet its learned features organize generators by
architectural similarity. The mechanism is direct: generators that
share components, such as base weights or an autoencoder, imprint
correlated artifacts on their images, so the classifier places those
images near one another in feature space; clustering the per-model
centroids then recovers their shared ancestry.
We extract penultimate-layer features ($1024$-dim) from the trained
CNN for each generator's test images and average them per model to
obtain one feature vector per generator. We reduce these vectors to
their top $64$ principal components, then apply unsupervised
hierarchical clustering (average linkage, correlation distance)
across the 27 OpenFake generators (Figure~\ref{fig:dendogram}).

The recovered hierarchy matches known lineage and faithfully
preserves the feature distances (cophenetic $r{=}0.910$). Seven of the eight Flux-family generators form one clade, including
HiDream (which reuses the Flux autoencoder) but not Flux~1.1~Pro; the
five SDXL variants group tightly, and the older Stable~Diffusion models
separate; SD~3.5, a from-scratch DiT with a new autoencoder, sits
apart. This structure emerges with no lineage supervision.

\subsection{Clustering Unseen Sources}
\label{sec:analysis_clustering}

A strong test of the learned representation is whether the fingerprint naturally organizes unseen images by their true sources. We extract penultimate features using the 17-class OpenFake model trained in \S\ref{sec:exp_openset}. With no labels and no retraining, we cluster images from the ten held-out generators (Figure~\ref{fig:discovery_umap}). Given no target count, density-based clustering auto-estimates seven clusters and no noise group against the true ten sources (ARI~0.64, NMI~0.84). Architecturally distinct generators (DALL$\cdot$E~3, Imagen~3, SDXL-EpicRealism) separate cleanly, and collapses occur only within a shared model family (e.g., Stable-Diffusion and Flux variants merging), again confirming that closer models yield more similar features (cf.\ \S\ref{sec:analysis_lineage}). The features thus effectively capture the shared structural artifacts of unseen generators, recovering both how many sources are present and which images share one. Full clustering metrics are in the supplementary.

\subsection{What Does the CNN See?}
\label{sec:analysis_whatseen}

The representation ablation of \S\ref{sec:exp_ablation} hints that
\method{} relies on subtle, non-semantic cues, which are washed out once
an image is encoded by a visual encoder. We deepen the analysis by
probing the spectral profile of the fingerprint, using the clean
$25$-class DRAGON model of \S\ref{sec:exp_blackbox}. We first ask which
frequencies suffice. We band-limit test images with a Butterworth filter
before attribution, sweeping the fraction of the spectrum removed, and
compare discarding the low frequencies against discarding the high ones
(Figure~\ref{fig:whatseen}, left). Accuracy survives the loss of the low
frequencies that carry most of the image content, and collapses
when the high frequencies go instead.  Using these findings, we design a
causal experiment that plants the fingerprint in images that never
carried it. From a host image $x_h$ and a generated
donor $x_d$, both split at $0.25$\,cyc/px, we build
\[
  \tilde{x} \;=\; \mathrm{LPF}(x_h) \;+\; \alpha\,\mathrm{HPF}(x_d),
\]
Here, $x_h$ is either a real photograph, an image from a generator other than the donor, or a flat gray field. We also construct a scrambled variant by partitioning $\mathrm{HPF}(x_d)$ into $8\times8$\,px blocks and randomly permuting their locations before transplantation. The scrambling is applied only to the donor’s high-frequency residue; $\mathrm{LPF}(x_h)$ remains unchanged.

As shown in Figure~\ref{fig:whatseen} (right), predictions increasingly follow the donor as $\alpha$ grows, across all three host types. This behavior persists after scrambling, albeit more weakly, indicating that most of the attribution signal does not depend on its global spatial arrangement. On generated hosts, the transplanted fingerprint also competes with and increasingly displaces the host’s native fingerprint. Examples of every construction and additional results are provided in the supplementary.

\subsection{Limitations}
\label{sec:analysis_limitations}
\paragraph{Generation settings are unexplored.}
DRAGON uses one known sampling configuration per generator, while the
configurations used in OpenFake are unknown. We therefore cannot determine
how much of the fingerprint comes from the model weights and how much from
choices such as guidance scale, step count, scheduler, or numerical precision.
Cross-benchmark transfer (\S\ref{sec:exp_adapt}) suggests some tolerance to
these differences, but controlled evaluation across configurations remains
future work.%

\paragraph{Adversarial robustness is untested.}
We evaluate robustness to common, non-adversarial corruptions (JPEG, blur,
resize, crop) but not to an adversary explicitly optimizing to evade the
classifier; we treat this as out of scope and flag it as a consequential
open task. Because \method{} is a standard image classifier, the extensive
literature on adversarial defenses (adversarial training, certified
defenses) applies directly and offers a concrete starting point.

\section{Conclusion}
\label{sec:conclusion}

We have demonstrated that a lightweight CNN operating on raw
image patches provides a powerful, scalable framework for model attribution,
achieving state-of-the-art accuracy in the strictest black-box settings.
Training for closed-set attribution also yields a versatile
fingerprint space that extends beyond its original objective and supports
various downstream tasks, including open-set rejection, few-shot adaptation
to new generators, unsupervised lineage recovery, and the grouping of unseen
images by source.

Ultimately, we hope this study demonstrates the power of simple, discriminative baselines over the heavy machinery now common in the field. Moreover, we aim to establish model attribution as a foundational problem in its own right—one where binary deepfake detection may eventually be viewed simply as a special case of source attribution.

\section*{Acknowledgement}
Part of this research was supported by ISF grant 2132/23.

\FloatBarrier
{
    \small  %
    \bibliographystyle{ieeenat_fullname}
    \bibliography{references}
}

\clearpage
\setcounter{section}{0}\setcounter{table}{0}\setcounter{figure}{0}
\renewcommand{\thesection}{S\arabic{section}}
\renewcommand{\thetable}{S\arabic{table}}
\renewcommand{\thefigure}{S\arabic{figure}}
\twocolumn[{\centering\Large\bf Supplementary Material\par\vspace{12pt}}]

This document provides supplementary material for the main paper:
architecture and training details (\S\ref{sec:training}), accuracy
and cost as a function of the patch budget (\S\ref{sec:efficiency})
and of the candidate count (\S\ref{sec:appendix_o1}), the
data-efficiency ablation (\S\ref{sec:dataeff}), closed-set and
cross-dataset confusion matrices (\S\ref{sec:closedconf} and \S\ref{sec:crossconf}), open-set and few-shot details
(\S\ref{sec:openset} and \S\ref{sec:fewshot}), what the
corruption-augmented model sees and examples of the transplant
constructions (\S\ref{sec:supp_freq} and \S\ref{sec:supp_transplant}),
and the lineage and unseen-source clustering metrics
(\S\ref{sec:lineage}).

\section{Architecture and Training Details}
\label{sec:training}

\paragraph{Architecture.}
The classifier operates directly on raw RGB patches, without an encoder:
each $64{\times}64{\times}3$ patch passes through three convolutional
blocks (a $3{\times}3$ stride-2 convolution, batch normalization, and
ReLU; channel widths $256$, $512$, $1024$).
A global average pool, dropout, and a linear head produce the class
logits, for $5.9$M parameters; only the output dimension $C$ (the
number of candidate generators) changes between experiments.

\paragraph{Training.}
We train end-to-end on labeled patches with class-balanced cross-entropy
and AdamW (learning rate $10^{-3}$ with cosine decay to $10^{-6}$, weight
decay $10^{-3}$), dropout $0.5$, and a batch of $256$ patches, for $75$
epochs with early stopping (patience $15$), keeping the checkpoint with
the best validation patch accuracy. Every patch inherits its image's
label and is weighted by $1/T$, $T$ being the image's tile count, so
that every image contributes equally regardless of resolution. The only
augmentation is random horizontal and vertical flips; inputs are
normalized by per-channel training-set statistics.

Two experiments modify this recipe. The corruption-augmented model is
trained on the union of seven copies of the training set: one clean and
six in which every image receives a single corruption, drawn once and
fixed for the run (JPEG at a quality from $\{60,70,80,90\}$; JPEG
$Q{=}60$; Gaussian blur $\sigma\!\in\![0.1,3.0]$ or $[3.0,4.0]$; resize
to $[384,768]$ or $[384,512]\,$px); corruptions are not composed. The
open-set backbones are trained with label smoothing of $0.2$, which
lowers known-class confidence and thereby widens the separation between
known and unknown confidence scores on which rejection is thresholded.

All latency and training-time figures in the main paper and in
\S\ref{sec:efficiency} were measured on a single NVIDIA GeForce
RTX~5090.

\section{Patch Budget and Inference Cost}
\label{sec:efficiency}

At test time an image is tiled into $64{\times}64$ patches (edge patches
overlap so that every pixel is covered) and the per-patch logits are
averaged into a single image-level prediction (Eq.~3 of the main paper).
The number of patches scored per image, the \emph{patch budget} $N$,
therefore controls the accuracy--compute trade-off at inference without
any change to the model.

Table~\ref{tab:efficiency} reports this trade-off; \emph{Fwd} is the
GPU forward pass alone, and \emph{Compute} adds patchification,
tensorization, and aggregation. Accuracy rises steeply and then
saturates: a single patch, $0.4\%$ of a $1024^2$ image, is sufficient
for most images, and sixteen patches come within $0.8$ points of the
full budget at a quarter of its cost. The reported times include the
surrounding pipeline; we believe they can be optimized further to
reflect the cost of the CNN forward pass itself.

\begin{table}[t]
\centering
\caption{%
    \textbf{Patch budget vs.\ accuracy, latency, and throughput.}
    Accuracy over all $25$ DRAGON generators; an image with fewer than
    $N$ tiles scores all of its tiles. Latency and throughput on
    $1024^2$ images, fp16, medians.
}
\label{tab:efficiency}
\small
\setlength{\tabcolsep}{4pt}
\begin{tabular}{@{}rcccc@{}}
\toprule
\textbf{$N$} & \textbf{Acc\,(\%)} & \textbf{Fwd (ms)} & \textbf{Compute (ms)} & \textbf{Thr.\,(img/s)} \\
\midrule
  1 & 91.94 & 0.13 & 0.70 & 1422 \\
  2 & 96.48 & 0.14 & 0.79 & 1259 \\
  4 & 97.56 & 0.16 & 0.71 & 1417 \\
  8 & 97.92 & 0.13 & 0.69 & 1450 \\
 16 & 98.17 & 0.20 & 0.75 & 1335 \\
 64 & 98.89 & 0.55 & 1.12 & 890 \\
256 & 98.94 & 2.04 & 2.73 & 367 \\
\bottomrule
\end{tabular}
\end{table}

\section{Inference Cost in the Candidate Count}
\label{sec:appendix_o1}

Adding a candidate generator widens the output layer, so inference cost
is not strictly constant in the candidate count $C$. A convolution with
a $k{\times}k$ kernel, $c_{\mathrm{in}}$ input channels,
$c_{\mathrm{out}}$ output channels, and an $H{\times}W$ output performs
$HW c_{\mathrm{out}} \cdot k^{2} c_{\mathrm{in}}$ multiply--accumulates,
two FLOPs each. Summed over the trunk's three stride-2 convolutions,
$32^{2}{\cdot}256{\cdot}3^{2}{\cdot}3 +
16^{2}{\cdot}512{\cdot}3^{2}{\cdot}256 +
8^{2}{\cdot}1024{\cdot}3^{2}{\cdot}512$ multiply--accumulates, the trunk
costs a fixed $1.22$\,GFLOP per $64^{2}$ patch independently of $C$; the
head is a single $1024 \times C$ matrix product costing $2048C$\,FLOP
per patch, so it reaches one percent of trunk cost only at
$C \approx 6{,}000$.
Table~\ref{tab:o1} confirms this on hardware with randomly initialized
networks of each width (the weights do not affect latency): latency is
essentially unchanged up to $C{=}10{,}000$ and rises measurably only
once the head dominates the parameter count. Parameter count and
compute diverge because the head is a single matrix product of fixed
depth: even where it holds most of the parameters, it contributes a far
smaller share of the arithmetic.

\begin{table}[t]
  \centering
  \caption{\textbf{Inference cost is nearly constant in the candidate count.}
    Latency is normalized to $C{=}1$.}
  \label{tab:o1}
  \small
  \setlength{\tabcolsep}{5pt}
  \begin{tabular}{@{}rrrrr@{}}
    \toprule
     & Latency & Params & \multicolumn{2}{c}{Head share (\%)} \\
    \cmidrule(l){4-5}
    $C$ & ($\times$\,$C{=}1$) & (M) & Params & FLOPs \\
    \midrule
          1     & 1.000 &   5.91 &  0.02 &    0.0002 \\
         10     & 1.002 &   5.92 &  0.17 &    0.0017 \\
        100     & 1.002 &   6.01 &  1.70 &    0.0168 \\
      1{,}000   & 1.006 &   6.94 & 14.8  &    0.1673 \\
     10{,}000   & 1.017 &  16.2  & 63.4  &    1.6482 \\
    100{,}000   & 1.154 & 108.4  & 94.6  &   14.3526 \\
    \bottomrule
  \end{tabular}
\end{table}

\section{Data Efficiency}
\label{sec:dataeff}

\begin{figure}[t]
  \centering
  \includegraphics[width=0.86\linewidth]{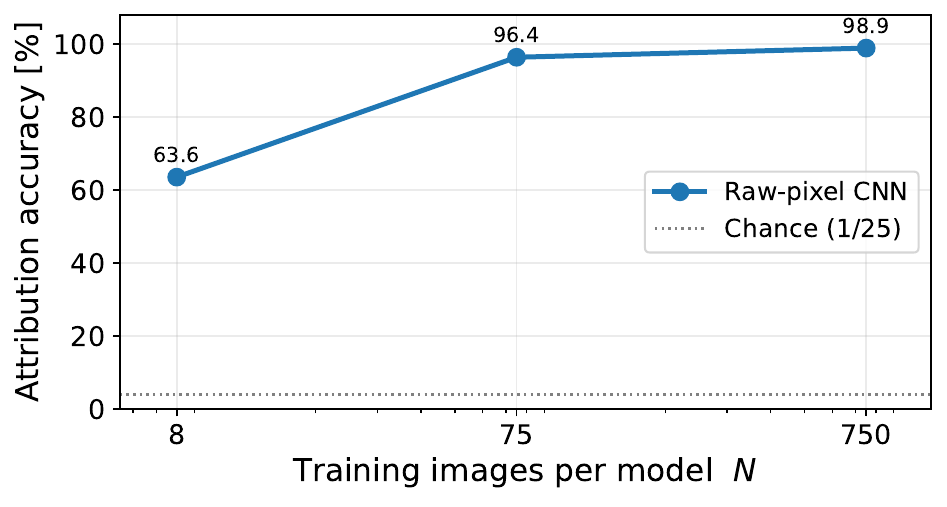}
  \caption{%
    Attribution accuracy vs.\ training images per generator on DRAGON
    ($25$ classes, log scale).}
  \label{fig:dataeff}
\end{figure}

We vary the number of training images per generator on DRAGON
($25$ classes) and score every model on the same $10{,}000$-image test
split (Figure~\ref{fig:dataeff}). Accuracy degrades gracefully: a
tenfold reduction from the full $750$ to $75$ images per generator
costs 2.5 points, and even $8$ images per generator support
attribution at 16 times the chance rate.

\section{Closed-Set Confusion Matrices}
\label{sec:closedconf}

\begin{figure*}[t]
  \centering
  \includegraphics[width=\linewidth]{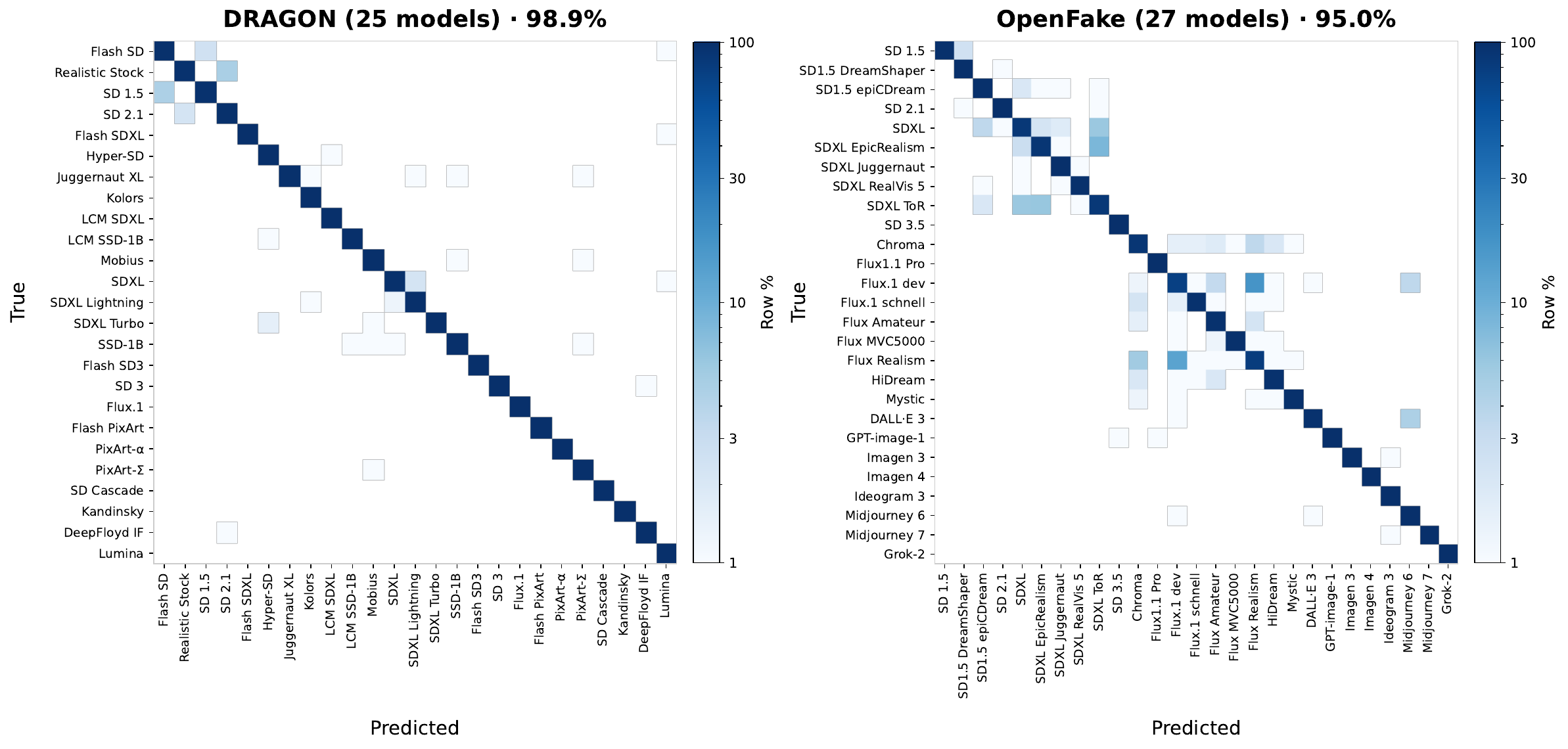}
  \caption{%
    Row-normalized confusion matrices for DRAGON (left) and
    OpenFake (right), ordered by family.}
  \label{fig:confusion}
\end{figure*}

Figure~\ref{fig:confusion} breaks the headline accuracies of the main
paper down into per-generator recall. Off-diagonal mass is concentrated
within model families, among the Flux variants and among the SDXL
fine-tunes: in both benchmarks about $85\%$ of the errors lie within a
family or between the adjacent SD~1.x and SD~2.x versions. Cross-family
errors are rare and unsystematic; on DRAGON, no generator pair
contributes more than three.

\section{Cross-Dataset Adaptation}
\label{sec:crossconf}

\begin{figure*}[t]
  \centering
  \includegraphics[width=\linewidth]{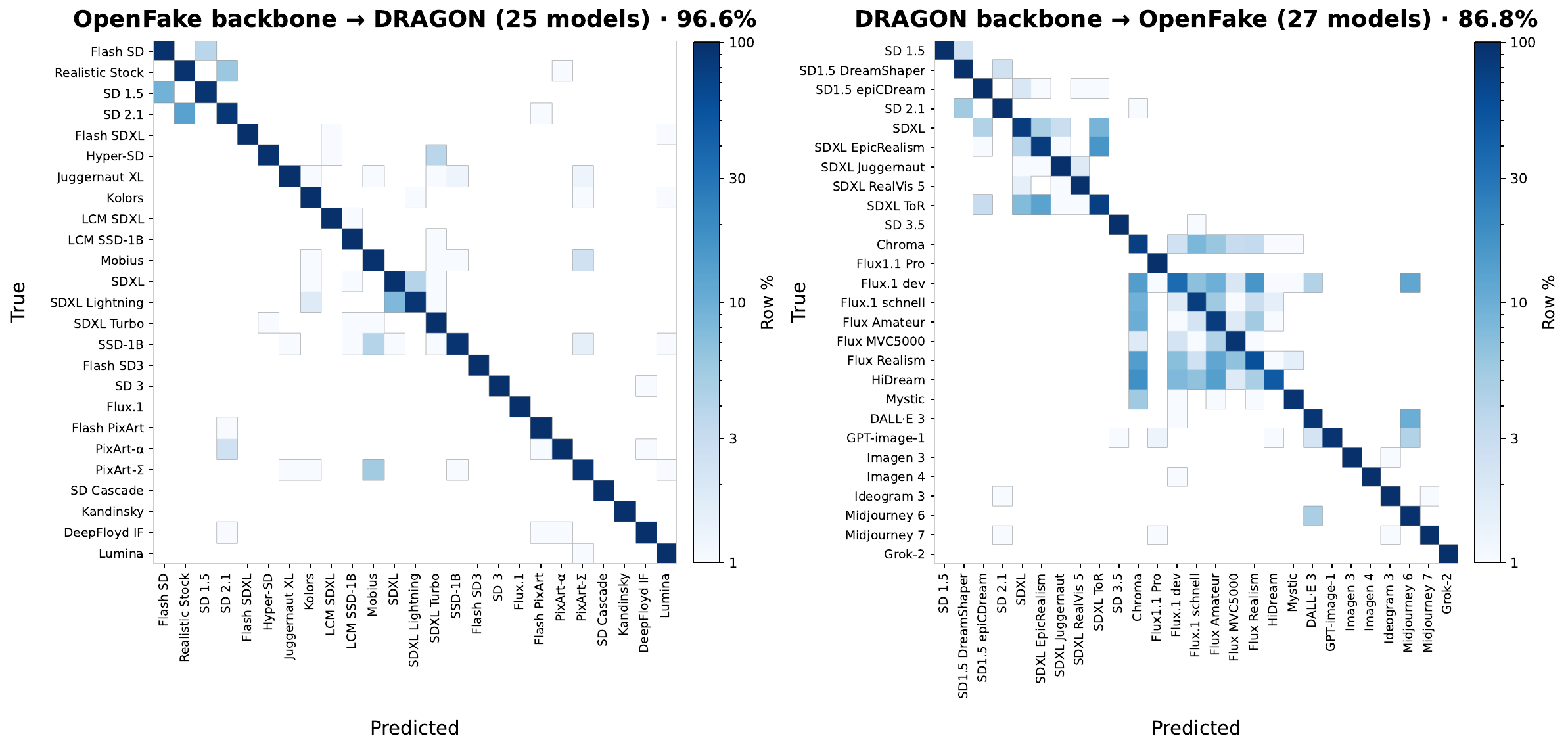}
  \caption{%
    Cross-dataset adaptation, one panel per direction, row-normalized
    per-generator recall (\%), ordered by family.}
  \label{fig:cm_cross}
\end{figure*}

The main paper transfers a backbone across benchmarks by freezing it and
fitting only a linear head on the full, disjoint generator set of the
other benchmark: an OpenFake backbone attributes DRAGON's $25$ generators
at $96.6\%$ ($98.9\%$ for the model trained on DRAGON), and a DRAGON
backbone attributes OpenFake's $27$ at $86.8\%$ ($95.0\%$ for the model
trained on OpenFake). Figure~\ref{fig:cm_cross} gives the per-generator
confusion matrices for both directions.

The OpenFake backbone transfers with little loss: $19$ of the $25$ DRAGON
generators are recalled at $95\%$ or above, and the errors are
concentrated on a few closely related pairs. In the reverse direction,
$83\%$ of the errors lie within a family, and the recall deficit is
concentrated on OpenFake's Flux and SDXL fine-tunes; DRAGON contains a
single Flux model, so a DRAGON backbone was never trained to separate
members of that family. Generators with a distinct architecture are
recalled at $99.5\%$ or above.

\section{Open-Set Attribution Details}
\label{sec:openset}

We randomly partition OpenFake's $27$ generators into $17$~\emph{known}
(used for training) and $10$~\emph{unknown} (held out entirely), and
repeat over five independent draws; reported scores are means over the
draws. An image is scored by the maximum softmax probability of its
patch-averaged logits and rejected as \emph{unknown} when that
confidence falls below a threshold $\tau$; no auxiliary outlier data
are used. The AU-OSCR and AUROC are threshold-free, and the
per-generator rejection rates set $\tau$ at the $5$th percentile of
known-generator scores ($95\%$ known-class true-positive rate).

\begin{figure}[t]
  \centering
  \includegraphics[width=0.86\linewidth]{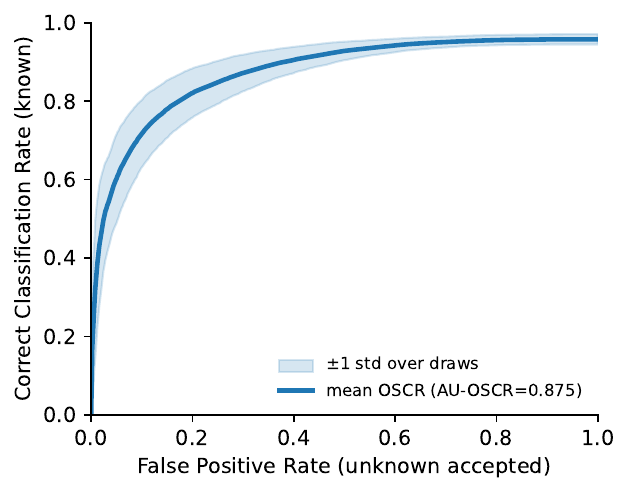}
  \caption{%
    Open-set classification rate (OSCR) curve, mean over the five draws
    with the $\pm1$ standard-deviation band.}
  \label{fig:oscr}
\end{figure}

Across the five draws the model attains an AU-OSCR of
$0.875 \pm 0.037$ (best draw $0.918$) and a known-versus-unknown AUROC
of $0.896 \pm 0.033$; Figure~\ref{fig:oscr} shows the OSCR curve, the
correct-classification rate on known generators against the
false-positive rate of accepting unknowns.

Table~\ref{tab:openset_reject} reports rejection per generator at the
$95\%$ known-TPR operating point, averaged over the draws in which the
generator was held out ($25$ of the $27$ are held out at least once). Fine-tunes of a family that remains among the known generators
are the most difficult to reject, the SDXL fine-tunes most of all, while
generators of a distinct lineage are rejected far more reliably. Three
generators largely evade rejection: Midjourney~6 (held out while
Midjourney~7 remains known), DALL$\cdot$E~3 and Grok-2 are assigned to a
known generator with high confidence.

\begin{table}[t]
  \centering
  \caption{Per-generator rejection at the $95\%$ known-TPR operating
    point, averaged over the held-out draws (\emph{Draws}). Flux MVC5000
    and SD1.5~epiCDream were never held out.}
  \label{tab:openset_reject}
  \small
  \setlength{\tabcolsep}{5pt}
  \begin{tabular}{llcc}
    \toprule
    Generator & Family & Draws & Reject (\%) \\
    \midrule
    Imagen 4          & Imagen     & 2 & 99 \\
    SD 3.5            & SD 3.x     & 1 & 96 \\
    SD1.5 DreamShaper & SD 1.x     & 1 & 88 \\
    Ideogram 3        & Ideogram   & 1 & 88 \\
    Flux 1.1 Pro      & Flux       & 2 & 84 \\
    SD 1.5            & SD 1.x     & 3 & 82 \\
    GPT Image 1       & GPT        & 2 & 77 \\
    Midjourney 7      & Midjourney & 2 & 69 \\
    SD 2.1            & SD 2.x     & 2 & 67 \\
    Imagen 3          & Imagen     & 2 & 64 \\
    Flux Realism      & Flux       & 3 & 55 \\
    HiDream           & Flux       & 1 & 55 \\
    Flux Amateur      & Flux       & 1 & 52 \\
    Mystic            & Flux       & 2 & 52 \\
    Flux.1 Dev        & Flux       & 3 & 49 \\
    Chroma            & Flux       & 4 & 49 \\
    Flux.1 Schnell    & Flux       & 3 & 46 \\
    SDXL              & SDXL       & 3 & 41 \\
    SDXL RealVis      & SDXL       & 1 & 37 \\
    SDXL EpicRealism  & SDXL       & 2 & 26 \\
    SDXL Juggernaut   & SDXL       & 2 & 18 \\
    Grok 2            & xAI        & 3 & 13 \\
    SDXL ToR          & SDXL       & 1 &  8 \\
    DALL$\cdot$E 3    & DALL-E     & 2 &  3 \\
    Midjourney 6      & Midjourney & 1 &  3 \\
    \bottomrule
  \end{tabular}
\end{table}

\section{Few-Shot Adaptation}
\label{sec:fewshot}

\begin{figure}[t]
  \centering
  \includegraphics[width=0.78\linewidth]{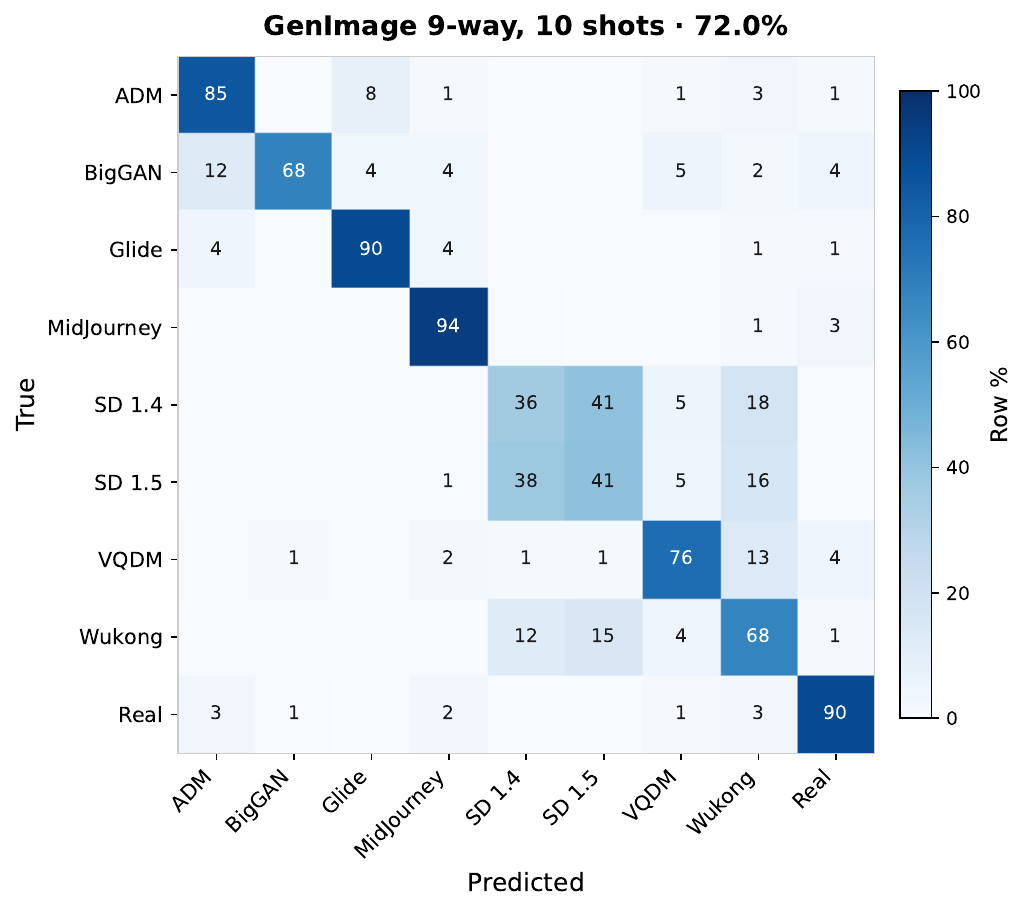}
  \caption{%
    GenImage $9$-class few-shot confusion (ten shots, frozen OpenFake
    backbone, row-normalized recall \%, pooled over five seeds).}
  \label{fig:fewshot}
\end{figure}

The few-shot experiment of the main paper freezes an OpenFake backbone
and fits a $9$-way linear head on GenImage from ten labeled images per
generator, reaching $72.0\%$. Figure~\ref{fig:fewshot} gives the
confusion matrix; errors are concentrated among SD~1.4, SD~1.5 and Wukong.

\section{What the Corruption-Augmented Model Sees}
\label{sec:supp_freq}

\begin{figure*}[!t]
  \centering
  \includegraphics[width=0.82\linewidth]{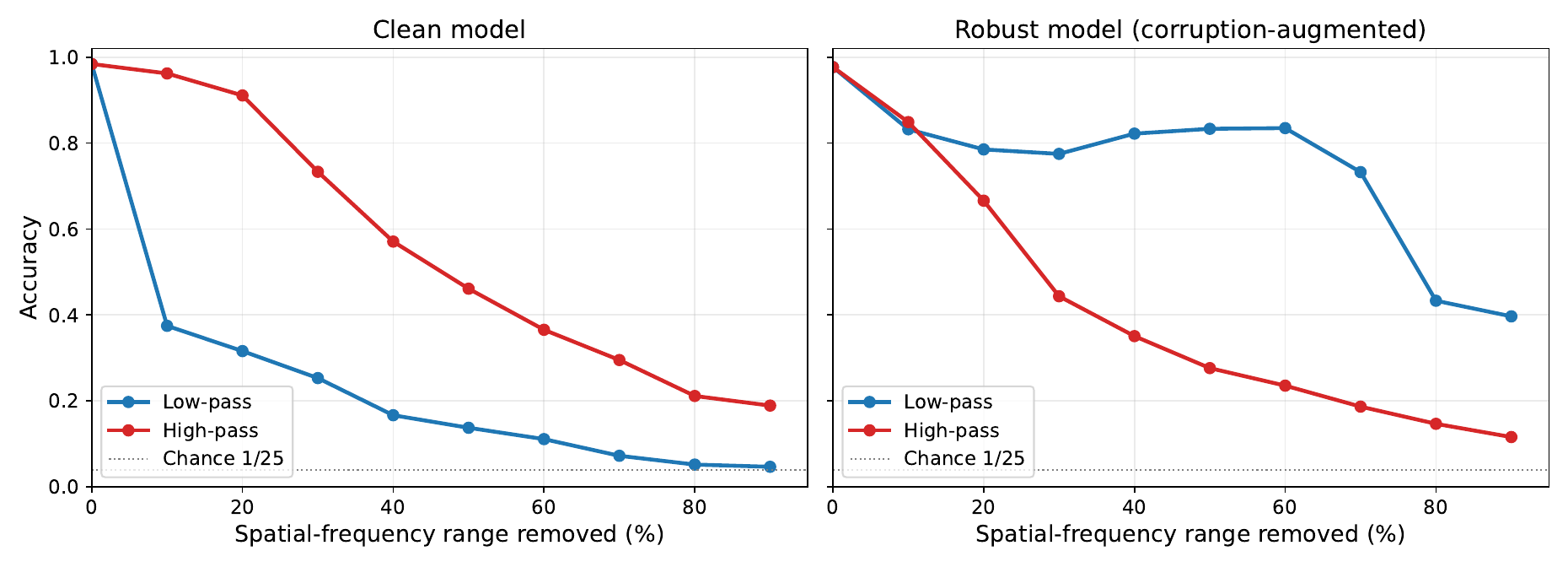}
  \caption{\textbf{Frequency reliance, clean vs.\ robust.} Attribution
    accuracy on DRAGON as low- or high-pass filtering removes part of the
    spectrum (Butterworth). \emph{Left:} clean model; \emph{right:}
    corruption-augmented model.}
  \label{fig:supp_freq}
\end{figure*}

\begin{figure*}[!b]
  \centering
  \includegraphics[width=0.78\linewidth]{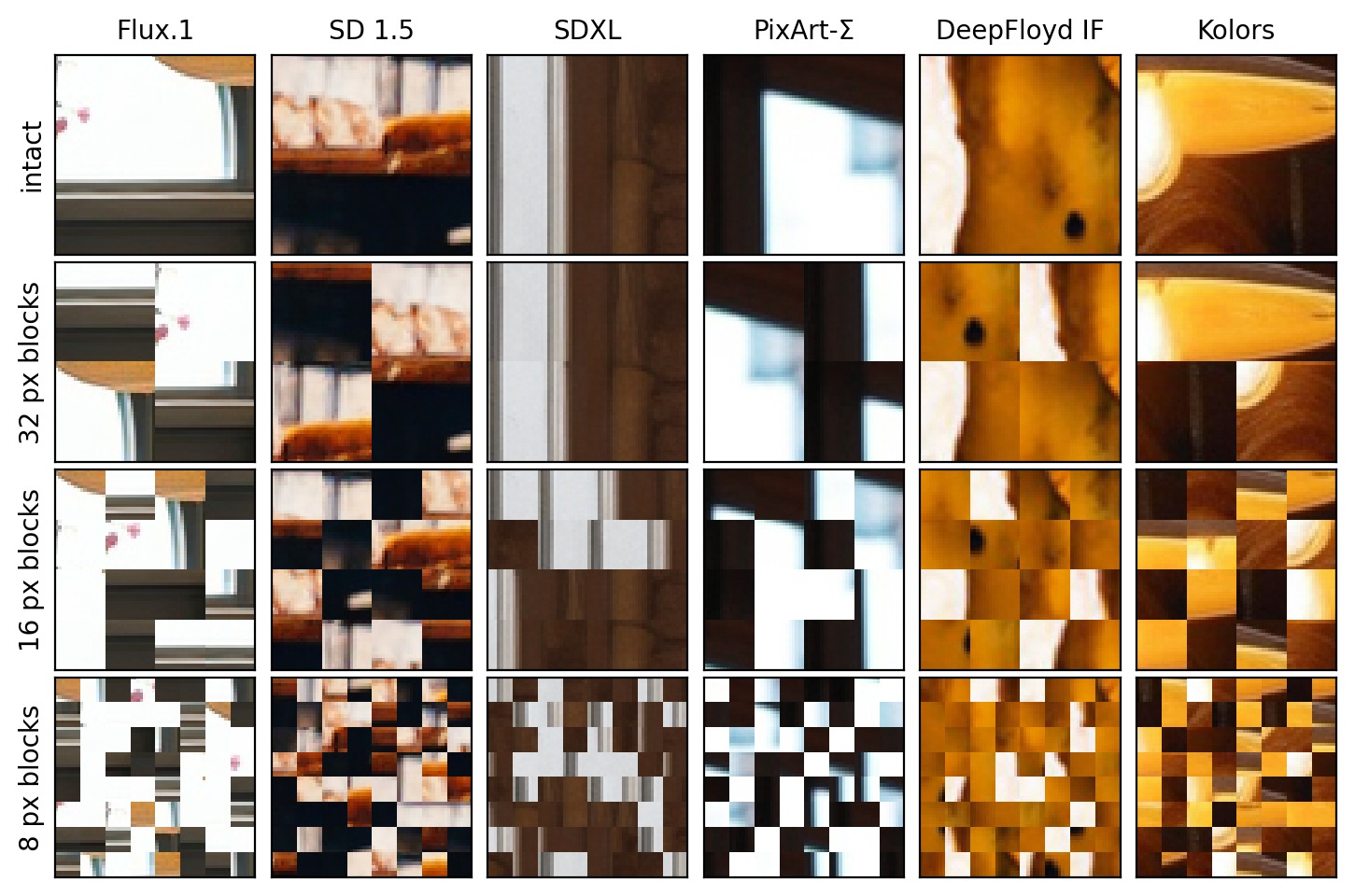}
  \caption{\textbf{Block scramble inside a $64$\,px patch.} One patch
    per generator (columns), intact and at three block sizes (rows).}
  \label{fig:supp_scramble}
\end{figure*}

The main paper band-limits test images with a Butterworth filter before
attribution and finds that the clean model reads the fingerprint from the
high spatial frequencies. We repeat that sweep on the corruption-augmented
model of the main paper's robustness experiment
(Figure~\ref{fig:supp_freq}); both are $25$-class DRAGON attributors
evaluated on the same test images. The augmented model shows the reverse
profile. With half of the spectral range removed, the clean model retains
$46.1\%$ accuracy when the low half is discarded and $13.8\%$ when the
high half is; the augmented model retains $27.6\%$ and $83.3\%$ under the
same two conditions, and $83.5\%$ with the upper $60\%$ of the range
removed.

This shift is consistent with the augmentation. JPEG compression and blur
both attenuate high frequencies, so a model trained under them is pushed
toward lower, more durable bands, which accounts for its graceful
degradation under those corruptions. The shift is partial: removing the
top tenth of the spectrum costs the augmented model $14.5$ points, against
$12.8$ for the bottom tenth, so it continues to exploit the high band
where that band is present.

Reliance on lower frequencies does not by itself imply reliance on
semantics, and two observations indicate that the augmented model's cue
remains low-level. First, the classifier operates on a single
$64{\times}64$ patch, $1/256$ of a $1024^2$ image. Second, the model is
largely insensitive to the spatial arrangement within that window. We scramble each $64$\,px patch by
partitioning it into $B{\times}B$ blocks and permuting the blocks,
independently per patch (Table~\ref{tab:scramble}). This destroys all
structure at scales above $B$ while preserving the statistics within each
block; the block boundaries introduce edges of their own, which add noise
rather than any generator-specific signal. Permuting whole patches across the image leaves every
prediction unchanged, as patches are scored independently and their logits
averaged. Within patches, the augmented model is nearly unaffected at
$B{=}32$ and remains far above chance at $B{=}16$, where the clean model
is the more tolerant; both models approach chance by $B{=}8$. Figure~\ref{fig:supp_scramble} illustrates
the operator: at $16$\,px no recognizable structure remains, only local
texture. Both models therefore rely on statistics confined to
neighborhoods of a few tens of pixels; the augmented model reads a coarser
band of that texture than the clean one.

\begin{table}[t]
  \centering
  \caption{Accuracy (\%) under within-patch block scrambling with block
    size $B$; chance is $4.0\%$, and $B{=}64$ is the unmodified patch.}
  \label{tab:scramble}
  \small
  \setlength{\tabcolsep}{4.5pt}
  \begin{tabular}{lcccccc}
    \toprule
    $B$ (px) & 64 & 32 & 16 & 8 & 4 & 2 \\
    \midrule
    Clean & 98.4 & 98.1 & 83.6 & 34.5 & 7.8 & 7.9 \\
    Corruption-augmented & 97.7 & 94.8 & 46.0 & 13.0 & 6.5 & 8.7 \\
    \bottomrule
  \end{tabular}
\end{table}

\section{Transplant Examples}
\label{sec:supp_transplant}

\begin{figure*}[!t]
  \centering
  \includegraphics[width=\linewidth]{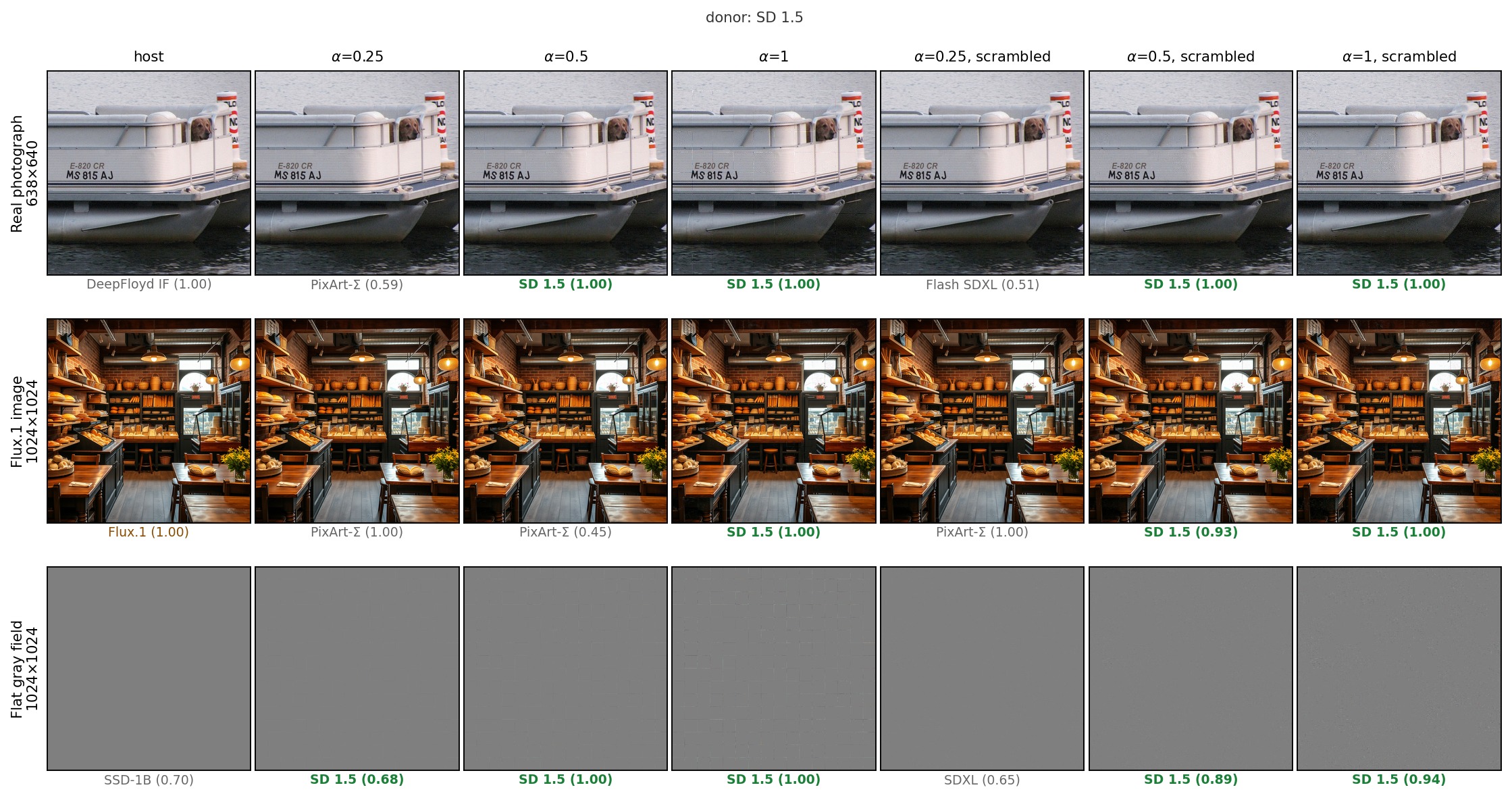}
  \caption{\textbf{Every construction of the transplant test.} Donor
    SD~1.5; rows are the three host types, columns the unmodified host,
    then three amplitudes $\alpha$ with the donor's high band intact and
    the same three with it scrambled into $8$\,px blocks. Labels give the
    model's prediction and confidence; predictions matching the donor are
    shown in green.}
  \label{fig:supp_transplant}
\end{figure*}

Figure~\ref{fig:supp_transplant} shows every construction of the causal
test for one donor, SD~1.5, on whole images at native resolution: the
low band of the host (a real photograph, a Flux.1 image, or a flat gray
field; Butterworth split at $0.25$\,cyc/px) is recombined with $\alpha$
times the donor's high band, assembled as in the experiment from
randomly drawn $64$\,px donor patches and either left intact or
scrambled into $8$\,px blocks. The model has no class for real images,
so the unmodified photograph and the gray field are necessarily assigned
to some generator.
At $\alpha{=}1$ every host is attributed to the donor, intact or
scrambled; at low amplitude the attenuated band is attributed to
PixArt-$\Sigma$. The switching amplitude varies across donors and
hosts, so the donor-averaged curves in the main paper rise more
gradually than this example.

\section{Lineage and Unseen-Source Clustering}
\label{sec:lineage}

\paragraph{Lineage.}
The main paper clusters the per-generator centroids of penultimate-layer
features ($27$ OpenFake generators, $64$ principal components, average
linkage on correlation distance); the tree preserves the feature
distances with cophenetic
$r{=}0.910$. Table~\ref{tab:lineage} compares linkage rules and gives
the bootstrap support of the tree, the number of internal branches that
recur when the test images are resampled and the tree rebuilt. HiDream
clusters with the
Flux family, whose autoencoder it reuses; SD~3.5, a from-scratch DiT with
a new autoencoder, is separated from the older Stable~Diffusion models;
same-family fine-tune pairs form the closest leaves.

\begin{table}[!t]
  \centering
  \caption{Linkage-rule comparison: cophenetic correlation with the
    feature distances, and the number of the $26$ internal branches
    supported by at least $95\%$ of $1{,}000$ bootstrap replicates.}
  \label{tab:lineage}
  \small
  \begin{tabular}{lcc}
    \toprule
    Linkage & Cophenetic $r$ & Branches $\geq 95\%$ \\
    \midrule
    Average \emph{(main paper)} & \textbf{0.910} & 21 / 26 \\
    Complete                    & 0.894 & --- \\
    Ward                        & 0.775 & --- \\
    \bottomrule
  \end{tabular}
\end{table}

\begin{table}[!b]
  \centering
  \caption{Unsupervised clustering of the ten held-out generators in the
    feature space of each draw's $17$-class model. The partitional
    algorithms run at the true $K{=}10$; HDBSCAN estimates $K$. Draw~5 is
    shown in the main paper.}
  \label{tab:clustering}
  \small
  \setlength{\tabcolsep}{3pt}
  \begin{tabular}{llcccccc}
    \toprule
     & & \multicolumn{5}{c}{Draw} & \\
    \cmidrule(lr){3-7}
    Algorithm & Metric & 1 & 2 & 3 & 4 & 5 & Mean \\
    \midrule
    HDBSCAN & ARI & 0.50 & 0.48 & 0.48 & 0.60 & 0.64 & 0.54 \\
     & NMI & 0.79 & 0.74 & 0.75 & 0.80 & 0.84 & 0.78 \\
     & Purity & 0.90 & 0.97 & 0.85 & 0.92 & 0.98 & 0.93 \\
     & Clusters & 9 & 5 & 9 & 8 & 7 & 7.6 \\
    \midrule
    Spectral & ARI & 0.52 & 0.51 & 0.49 & 0.63 & 0.69 & 0.57 \\
     & NMI & 0.74 & 0.70 & 0.71 & 0.79 & 0.83 & 0.76 \\
     & Purity & 0.80 & 0.68 & 0.76 & 0.87 & 0.84 & 0.79 \\
     & Clusters & 10 & 10 & 10 & 10 & 10 & 10.0 \\
    \midrule
    K-means & ARI & 0.50 & 0.47 & 0.54 & 0.59 & 0.67 & 0.55 \\
     & NMI & 0.67 & 0.64 & 0.68 & 0.73 & 0.80 & 0.70 \\
     & Purity & 0.69 & 0.65 & 0.71 & 0.72 & 0.83 & 0.72 \\
     & Clusters & 10 & 10 & 10 & 10 & 10 & 10.0 \\
    \midrule
    Agglomerative & ARI & 0.33 & 0.44 & 0.39 & 0.51 & 0.51 & 0.44 \\
     & NMI & 0.60 & 0.68 & 0.70 & 0.73 & 0.72 & 0.69 \\
     & Purity & 0.90 & 0.94 & 0.95 & 0.90 & 0.92 & 0.92 \\
     & Clusters & 10 & 10 & 10 & 10 & 10 & 10.0 \\
    \bottomrule
  \end{tabular}
\end{table}

\paragraph{Clustering unseen sources.}
The main paper extracts penultimate features with a $17$-class model for
the test images of its ten held-out generators and clusters them without
labels or retraining. Table~\ref{tab:clustering} gives the metrics for
all five draws of \S\ref{sec:openset}: HDBSCAN on a UMAP embedding with
the cluster count estimated automatically (the main paper's setting),
and three partitional algorithms at the true $K{=}10$.
HDBSCAN returns five to nine clusters at high purity: nearly every
cluster is dominated by a single generator, and the ARI deficit arises
from within-family merges rather than splits. Draw~5, shown in the main
paper, is the best of the five; the partitional algorithms reach
comparable NMI at lower purity.

\end{document}